\documentclass[10pt,conference]{ieeeconf}

\IEEEoverridecommandlockouts

\usepackage{cite}
\usepackage{amsmath,amssymb,amsfonts,mathtools}
\usepackage{algorithmic}
\usepackage{algorithm}
\usepackage{graphicx}
\usepackage{textcomp}
\usepackage{xcolor}
\usepackage{tikz}
\usepackage[caption=false,font=footnotesize]{subfig}

\def\BibTeX{{\rm B\kern-.05em{\sc i\kern-.025em b}\kern-.08em
    T\kern-.1667em\lower.7ex\hbox{E}\kern-.125emX}}

\title{Freehand Sketching for End-User Programming of Robot Swarms}

\author{
    Riwa Karam$^1$, Ian Kuo$^1$, and Magnus Egerstedt$^2$
    \thanks{
        $^1$R. Karam and I. Kuo are with the Samueli School of Engineering, University of California, Irvine, Irvine, CA, 92697, USA. Email: {\tt\small \{rwkaram, ikuo3\}@uci.edu}.
    }
    \thanks{
        $^2$M. Egerstedt is with the University of North Carolina at Chapel Hill, Chapel Hill, NC, 27599, USA. Email: {\tt\small \{magnus\}@unc.edu}.
    }
}

\begin{document}

\maketitle

\begin{abstract}
    Robot swarms are increasingly used in applications where accessible interaction with non-expert users is desirable. This paper investigates freehand sketching as an end-user programming interface for specifying robot swarm geometries. Users communicate spatial intent through a drawing, while the swarm autonomously extracts target formation points, constructs a rigid formation graph, assigns robots to formation nodes, and executes distributed formation control with a guarantee against unintended reflected formations. The resulting sketch-to-swarm framework is evaluated through a human study examining the usability of freehand formation specification. Twenty participants generated $42$ geometric shapes, and the interface achieved a mean System Usability Scale score of $84.25$, which conventionally indicates high perceived usability. The results support freehand sketching as an intuitive interaction abstraction for human-swarm collaboration without requiring robotics or programming expertise.
\end{abstract}

\section{Introduction} \label{sec:introduction}

Robot swarms enable large groups of robots to accomplish tasks through decentralized coordination, collective perception, and distributed decision making~\cite{brambilla2013swarm,dorigo2021swarm}. Their scalability and robustness have motivated applications in domains such as environmental monitoring, exploration, search and rescue, precision agriculture, and logistics. More recently, robot swarms have also found use in educational platforms, public demonstrations, and artistic performances, where the spatial organization of the robots becomes an expressive medium rather than merely a functional one~\cite{mclurkin2012using,huang2021unmanned,serpiva2021swarmpaint}. As robot swarms become increasingly visible outside traditional robotics applications, an important challenge is therefore not only how robots coordinate with one another, but how humans, particularly users without expertise in robotics, control, or programming, can contribute their intent to the multi-robot system.

Human-Swarm Interaction (HSI)~\cite{kolling2015human} provides a natural setting for human-robot collaboration, in which complementary human and robotic capabilities are combined toward outcomes that neither can achieve independently~\cite{karam2026collaboration}. In such settings, the human typically contributes high-level intent and task-specific knowledge, while the swarm provides distributed coordination, autonomous decision making, and physical execution. Realizing this complementarity requires interaction abstractions that enable non-expert users to communicate collective intent without reasoning about resource or task assignments, swarm dynamics, or low-level control parameters.

Existing HSI interfaces have explored a variety of such interaction abstractions, including mixed-granularity interaction~\cite{patel2019mixed}, gesture-based control~\cite{macchini2021personalized,serpiva2021swarmpaint}, augmented and virtual reality~\cite{jang2021omnipotent,sachidanandam2022effectiveness}, and scale-invariant spatial specifications~\cite{meyer2023scale}. These approaches demonstrate that appropriately designed interfaces can reduce the cognitive and technical burden associated with commanding large multi-robot systems. However, they primarily address navigation, trajectory generation, task allocation, or coverage, where users specify how a swarm should move or behave. Less attention has been devoted to interaction paradigms through which a user can directly specify \emph{what} geometric shape a swarm should collectively realize without selecting from predefined commands, gestures, or geometries.

Enabling users to specify desired swarm geometries directly also aligns with end-user robot programming, which investigates methods for specifying robot behaviors without conventional programming~\cite{ajaykumar2021survey}. Proposed approaches include visual programming, programming by demonstration, tangible interaction, and multi-modal authoring, in which higher-level user specifications are translated into executable robot behaviors~\cite{porfirio2021figaro,porfirio2023sketching}. Extending this paradigm to robot swarms motivates representations through which users can directly specify collective behaviors while delegating their realization to the swarm. Freehand sketching provides a high-level spatial abstraction for this purpose. Rather than explicitly programming the motion of individual robots, the human contributes the desired geometric intent through a drawing.

Prior work established the foundations for transforming freehand drawings into robot formations through computer vision, rigid-graph construction, optimal robot assignment, and distributed formation control, together with theoretical guarantees preventing unintended reflected realizations of asymmetric formations~\cite{karam2025graphical}. While this work established the computational feasibility of translating sketches into robot-swarm formations, evaluating the efficacy of freehand sketching as an interaction abstraction for end-users remains a gap, particularly for users without prior robotics or control experience.

This paper addresses this gap by investigating freehand sketching as an end-user programming interface for collaborative human-swarm formation specification. Building upon the aforementioned prior work, human spatial specification and autonomous swarm realization are integrated into an interactive sketch-to-swarm system and evaluated through a human study. The study examines the perceived usability of the interface and evaluates whether users can interact with the complete sketch-to-swarm pipeline on a physical multi-robot platform.

The main contributions of this paper are:
\begin{itemize}
    \item A sketch-based interaction abstraction for human-swarm collaboration in which non-expert users communicate formation-level intent through freehand drawings without requiring conventional programming expertise or a set of predefined formation shapes.
    \item A human study with 20 participants evaluating the perceived usability and learnability of sketch-based formation specification.
    \item Physical Robotarium \cite{pickem2017robotarium} experiments across 42 self-specified participant formations, all of which converged to the corresponding user-specified formation.
\end{itemize}

\section{Sketch-to-Swarm Formation Framework} \label{sec:framework}

In this section, the sketch-to-swarm formation framework is described. The user specifies the desired formation shape through the sketch, while geometric interpretation, robot assignment, and robot-level control are handled by the system. We first elaborate on the processing of the sketch and how the formation is constructed, followed by formalizing the robot-to-formation assignment problem, and then presenting the reflection-free result for the distributed formation execution.

\subsection{Sketch Processing and Formation Construction} \label{ssec:sketch_interp}

\begin{figure}[!t]
    \centering
    \subfloat[Sketch-Processing Pipeline.]{
        \includegraphics[width=0.8\columnwidth]{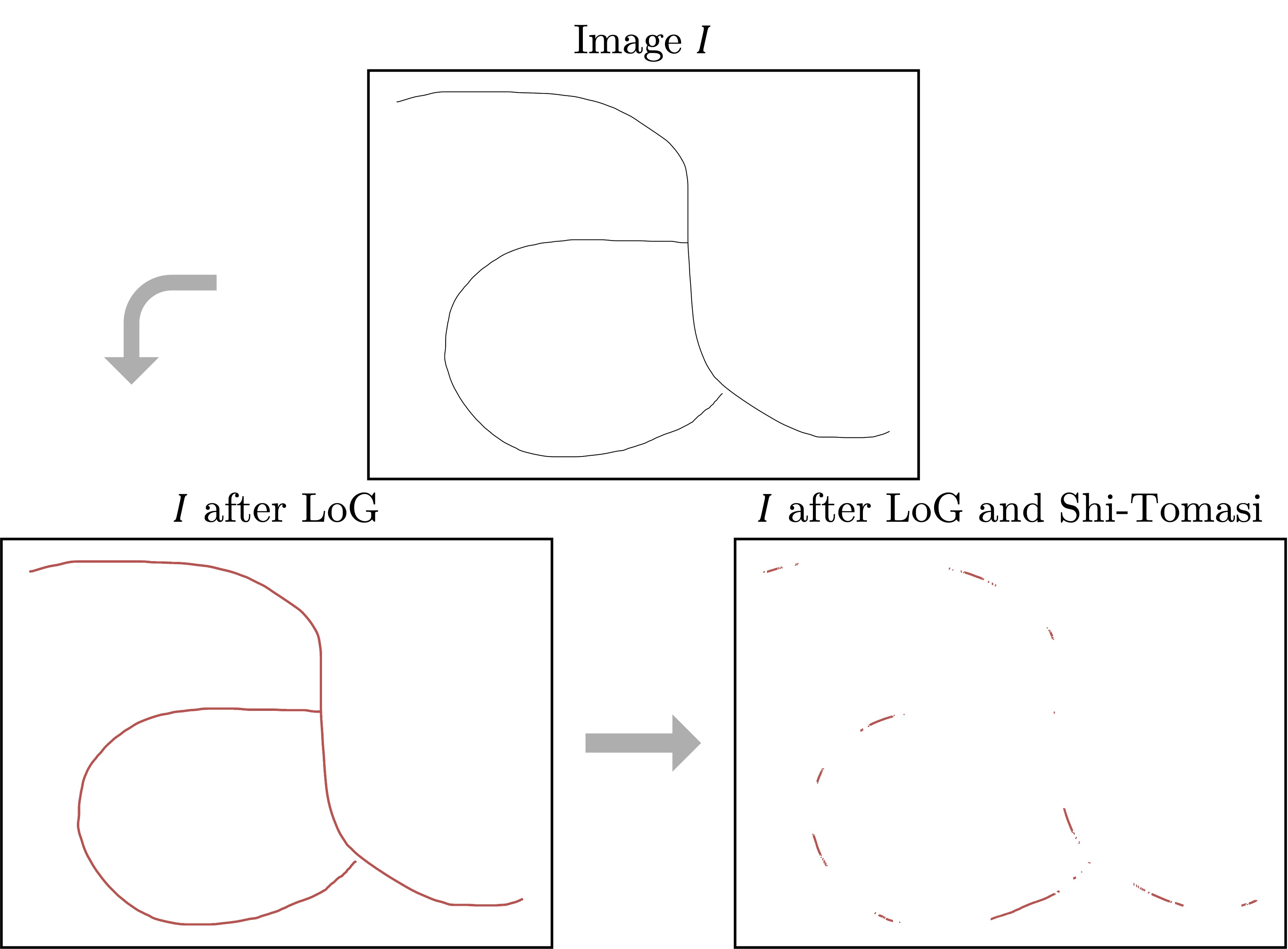}
        \label{fig:letter_a_CV}
    }
    \hfill
    \subfloat[Scaled and Filtered Target Points (x- and y-axes in meters).]{
        \includegraphics[width=0.85\columnwidth]{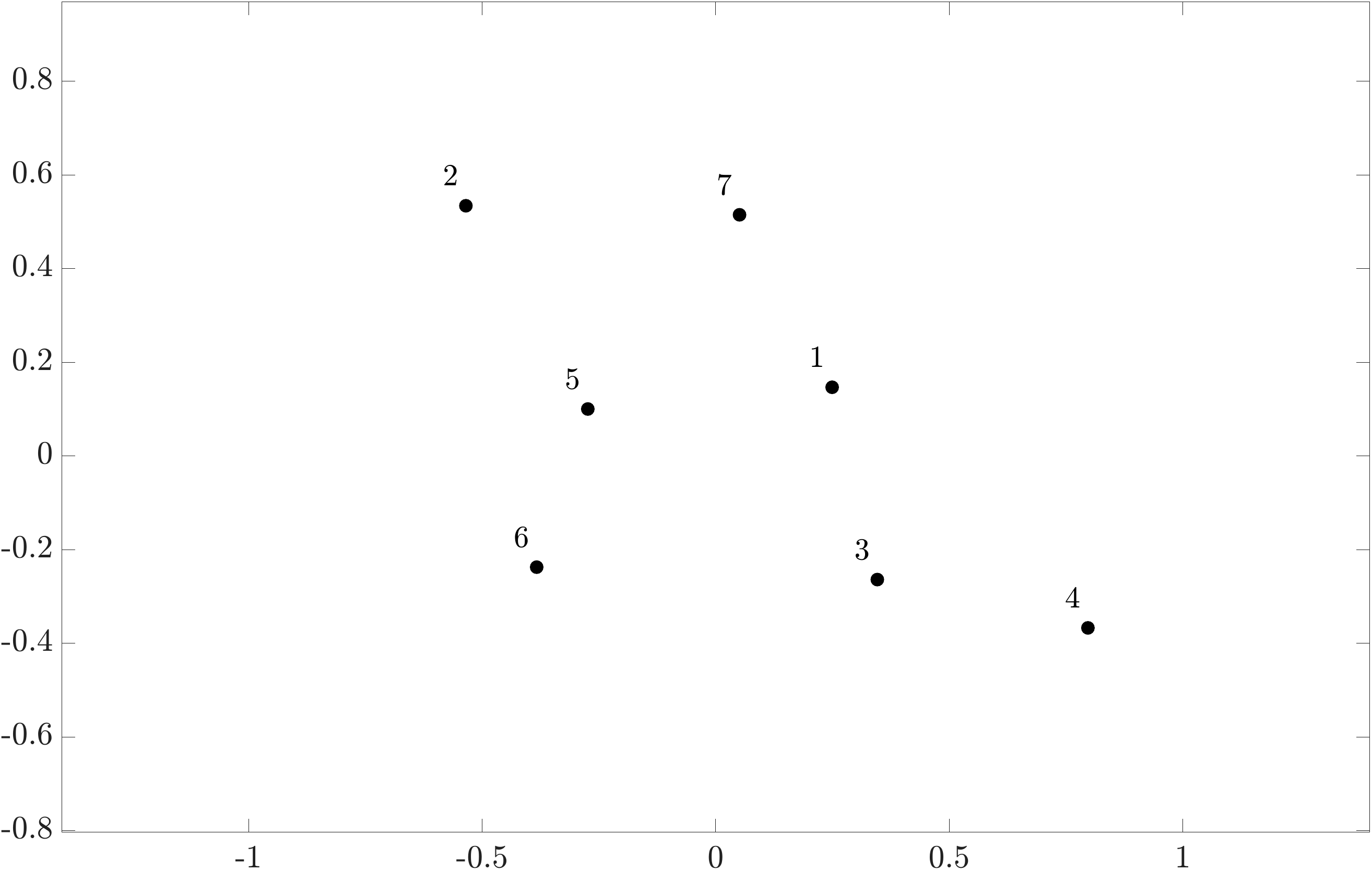}
        \label{fig:critical_points}
    }
    \caption{Sketch processing for an example user-drawn letter ``a''. (a) The input image $I$ is processed using Laplacian-of-Gaussian edge detection followed by Shi-Tomasi feature detection to identify important locations along the sketch. (b) The extracted points are scaled to the robot workspace and filtered according to the minimum-separation constraint, in this case $d_{\min}=0.35$m, and the algorithm produced 7 target points defining the desired formation geometry.}
    \label{fig:letter_a_CV_pipeline}
\end{figure}

Let $I$ denote the image generated from a user’s freehand sketch. To translate the sketch into a geometric representation suitable for formation specification. First, we use Laplacian of Gaussian (LoG)~\cite{marr1980theory}, an edge-detection method, to extract the drawing contours while reducing sensitivity to small variations in the freehand input. Shi-Tomasi feature detection~\cite{shi1994good} is a corner-detection method that identifies image locations exhibiting strong intensity variation in multiple directions. Hence, we then apply it to identify critical points along these contours, providing a compact set of candidate formation nodes that captures the sketch geometry. The extracted points are subsequently scaled from image coordinates to a physical robot workspace and filtered according to a minimum separation distance $d_{\min}$, ensuring that the specified geometry is compatible with the spatial and safety constraints of the physical robots. Figure~\ref{fig:letter_a_CV_pipeline} illustrates this process for an example freehand sketch of the letter ``a'', from the original user input through contour and feature extraction to the final set of scaled and filtered formation points.

After processing, the sketch is represented by $n$ target points $\mathbf{q}_i\in\mathbb{R}^2$, collected in the target configuration matrix
\begin{equation}
    Q =
    \begin{bmatrix}
        \mathbf{q}_1^T \\
        \mathbf{q}_2^T \\
        \vdots \\
        \mathbf{q}_n^T
    \end{bmatrix}
    \in\mathbb{R}^{n\times 2}.
    \label{eq:target_configuration}
\end{equation}
The number of extracted target points is constrained by the number of robots $N$ available for deployment, with $n \in [1,N]$. The target points are used to construct a formation graph
\begin{equation}
    \mathcal{G}=(\mathcal{V},\mathcal{E}),
\end{equation}
where each node $v_i\in\mathcal{V}$ corresponds to the index of a particular target point $\mathbf{q}_i$, and each edge $(v_i,v_j)\in\mathcal{E}$ is associated with a desired inter-agent distance defined by the corresponding target points
\begin{equation}
    d_{ij}=\|\mathbf{q}_i-\mathbf{q}_j\|_2.
    \label{eq:desired_distance}
\end{equation}

The formation graph is constructed to be rigid so that its prescribed edge distances determine the desired formation geometry up to rigid transformations~\cite{olfati2002graph,laman1970graphs}. Let $R\in\mathbb{R}^{|\mathcal{E}|\times 2n}$ denote the rigidity matrix associated with the framework $(\mathcal{G},Q)$, where each row corresponds to an edge $(v_i,v_j)\in\mathcal{E}$ and captures the dependence of the corresponding distance constraint on the target positions $\mathbf{q}_i$ and $\mathbf{q}_j$. In $\mathbb{R}^2$, the framework is infinitesimally rigid when
\begin{equation}
    \operatorname{rank}(R)=2n-3.
    \label{eq:rigidity_rank}
\end{equation}
Consequently, satisfaction of the prescribed edge-distance constraints locally preserves the formation shape while allowing global translation and rotation. Since pairwise distances are preserved under reflection, the same distance constraints can admit a mirrored configuration, motivating the reflection-free result discussed in Section~\ref{ssec:formation_control}.

\subsection{Robot-to-Formation Assignment} \label{ssec:assignment}

The extracted formation nodes specify the desired geometry but do not determine which physical robot should occupy each node. Let $\mathbf{p}_j\in\mathbb{R}^2$ denote the position of robot $j$. The robot positions are collected in the configuration matrix
\begin{equation}
    P =
    \begin{bmatrix}
        \mathbf{p}_1^T \\
        \mathbf{p}_2^T \\
        \vdots \\
        \mathbf{p}_n^T
    \end{bmatrix}
    \in\mathbb{R}^{n\times 2}.
    \label{eq:robot_configuration}
\end{equation}
The cost of assigning robot $j$ to target point $i$ is defined by their Euclidean distance,
\begin{equation}
    c_{ij}=\|\mathbf{p}_j-\mathbf{q}_i\|_2,
    \label{eq:assignment_cost}
\end{equation}
where $C=[c_{ij}]\in\mathbb{R}^{n\times n}$ is the assignment cost matrix.

Let $x_{ij}\in\{0,1\}$ be a binary assignment variable such that $x_{ij}=1$ if robot $j$ is assigned to formation node $i$, and $x_{ij}=0$ otherwise. The assignment variables are collected in the assignment matrix
\begin{equation}
    X=[x_{ij}]\in\{0,1\}^{n\times n}.
    \label{eq:assignment_matrix}
\end{equation}
The robot-to-node allocation is formulated as the binary linear assignment problem
\begin{equation}
    \begin{aligned}
        \min_{X} \quad &
        \sum_{i=1}^{n}\sum_{j=1}^{n} c_{ij}x_{ij} \\
        \text{s.t.}\quad &
        \sum_{j=1}^{n}x_{ij}=1,
        \qquad i=1,\ldots,n,\\
        &
        \sum_{i=1}^{n}x_{ij}=1,
        \qquad j=1,\ldots,n,\\
        &
        x_{ij}\in\{0,1\},
        \qquad i,j=1,\ldots,n.
    \end{aligned}
    \label{eq:assignment}
\end{equation}
The first constraint assigns exactly one robot to each formation node, while the second assigns each robot to exactly one node. Since the resulting optimization is a linear one-to-one assignment problem, the Hungarian algorithm~\cite{kuhn1955hungarian,munkres1957algorithms} is used to efficiently obtain a globally optimal assignment, minimizing the total robot-to-target distance.

\subsection{Reflection-Free Distributed Formation Execution} \label{ssec:formation_control}

Following the assignment, each robot is associated with one node of the desired formation graph and its corresponding target point. For notational simplicity, robot indices are relabeled according to this assignment such that robot $i$ corresponds to target point $\mathbf{q}_i$. The desired distance between neighboring robots $i$ and $j$ is therefore $d_{ij}$ as defined in Equation~\eqref{eq:desired_distance}. Since the desired formation is specified by the inter-agent distances associated with the rigid formation graph, a distance-based formation controller provides a natural distributed mechanism for realizing these constraints. Each robot moves according to the control law~\cite{olfati2002graph}
\begin{equation}
    \dot{\mathbf{p}}_i =
    -k\sum_{j\in\mathcal{N}_i}
    \left(
        \|\mathbf{p}_i-\mathbf{p}_j\|_2^2-d_{ij}^2
    \right)
    (\mathbf{p}_i-\mathbf{p}_j),
    \label{eq:formation_control}
\end{equation}
where $\mathcal{N}_i$ denotes the set of neighbors of robot $i$ in $\mathcal{G}$ and $k>0$ is a control gain. The controller drives the inter-agent distance errors toward zero using only relative information associated with neighboring robots, thereby realizing the geometry extracted from the user’s sketch in a distributed manner.

\begin{figure}[!t]
    \centering
    \subfloat[Horizontal Reflection.]{
        \includegraphics[width=0.47\columnwidth]{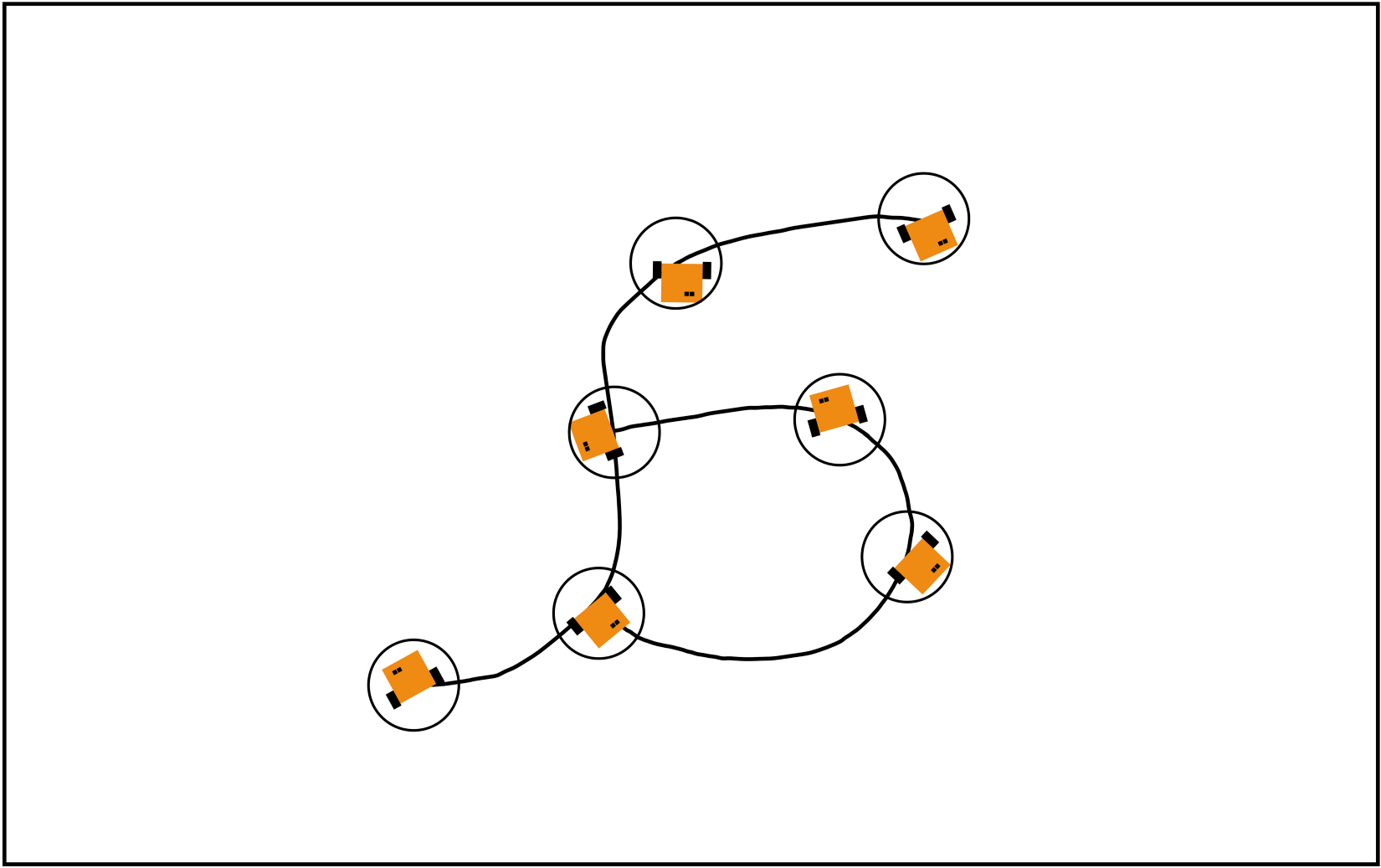}
        \label{fig:horizontal_flip}
    }
    \hfill
    \subfloat[Vertical Reflection.]{
        \includegraphics[width=0.47\columnwidth]{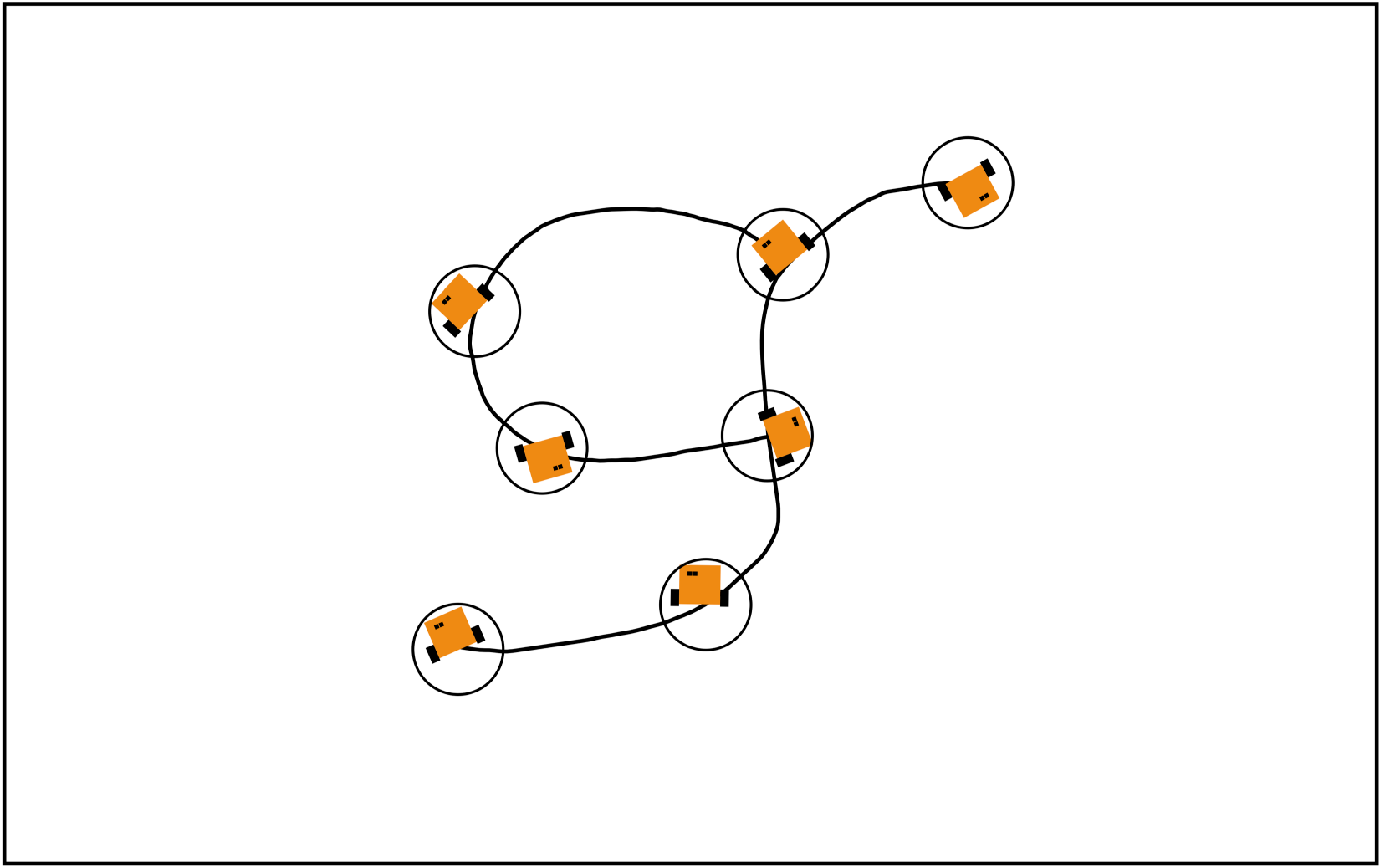}
        \label{fig:vertical_flip}
    }
    \caption{Examples of unintended reflected realizations of the asymmetric  letter ``a'' formation. Because pairwise distance constraints are invariant under reflection, both (a) a horizontal reflection and (b) a vertical reflection can satisfy the same inter-agent distances as the intended configuration while changing the orientation and semantic appearance of the user's sketch. The assignment stage is used to exclude such possible reflections.}
    \label{fig:flip_example}
\end{figure}

Because the controller in Equation~\eqref{eq:formation_control} regulates inter-agent distances rather than absolute robot positions, configurations related by translation, rotation, or reflections satisfy the same set of distance constraints. Translation and rotation preserve the intended appearance of the sketch, whereas reflection can alter the semantic content of an asymmetric formation. For example, a reflected letter or asymmetric symbol may no longer represent the configuration specified by the user.     Figure~\ref{fig:flip_example} illustrates this ambiguity for the asymmetric letter ``a’’ example.

\begin{figure}[!b]
    \centering
    \includegraphics[width=0.9\linewidth]{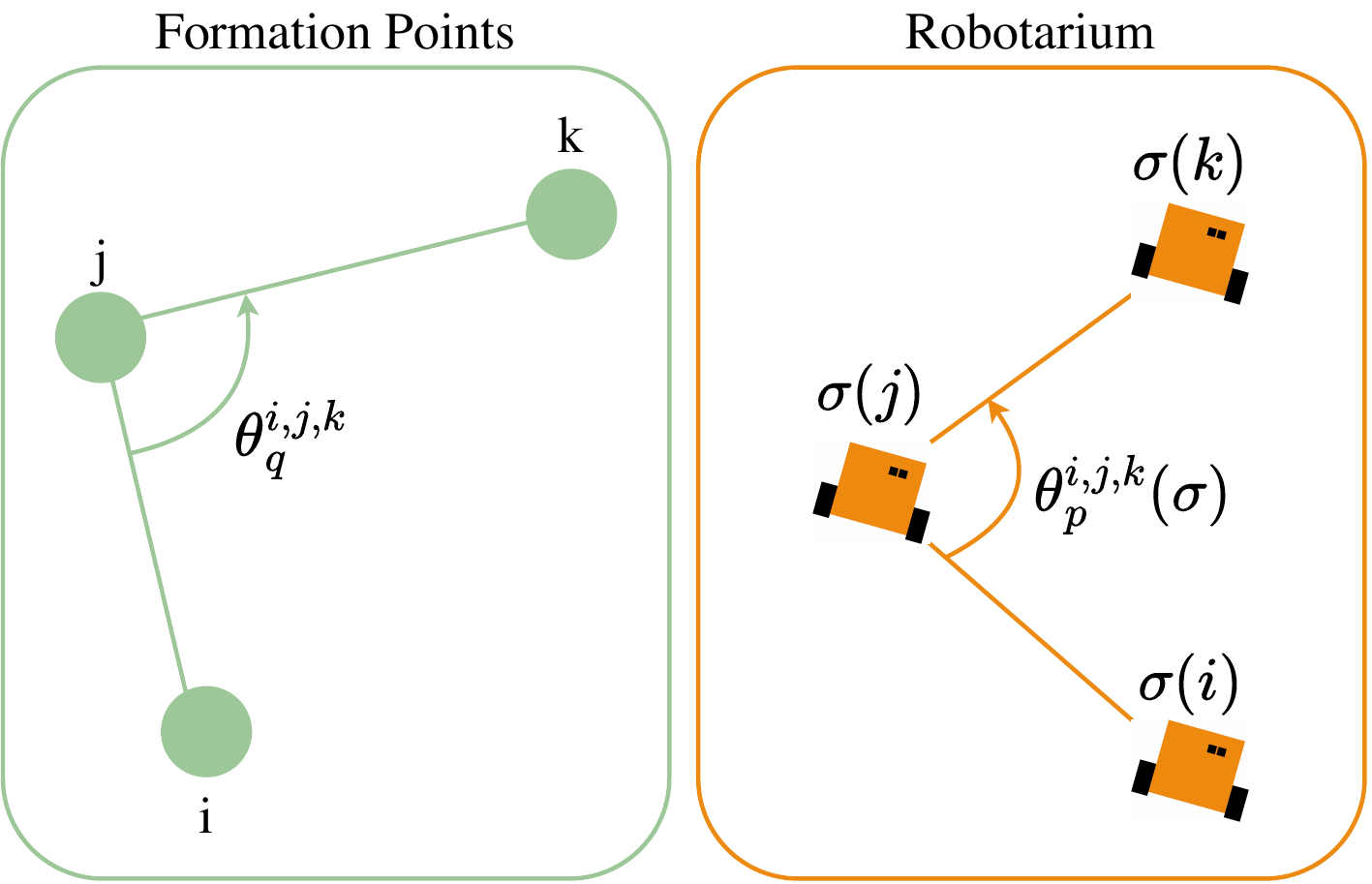}
    \caption{Illustration of the signed orientation angles used in the reflection-free condition. The target formation points indexed by $i$, $j$, and $k$ define the orientation angle $\theta_{q}^{i,j,k}$, while the robots assigned through the permutation $\sigma$ define the corresponding angle $\theta_{p}^{i,j,k}(\sigma)$. Preservation of the sign of these angles prevents the assigned robot configuration from being a reflected realization of the target configuration.}
    \label{fig:angles}
\end{figure}

To address this ambiguity, the assignment process is accompanied by a reflection-free result. Let $\sigma$ denote the minimum-cost permutation mapping target formation nodes to robots. For fixed robot and target configurations $P$ and $Q$, respectively, the corresponding squared assignment cost is defined as
\begin{equation}
    \phi(\sigma,P,Q)
    =
    \sum_{i=1}^{n}
    \|\mathbf{p}_{\sigma(i)}-\mathbf{q}_i\|_2^2.
    \label{eq:permutation_cost}
\end{equation}
For any three non-collinear formation points $i,j,k$, let $\theta_{q}^{i,j,k}$ denote their signed orientation angle and $\theta_{p}^{i,j,k}(\sigma)$ denote the corresponding signed orientation angle of the assigned robots. These angles are illustrated in Figure~\ref{fig:angles}, where the permutation $\sigma$ associates the target-point indices with their corresponding robots. The reflection-free result established in~\cite{karam2025graphical} states that, under the minimum-cost assignment,
\begin{equation}
    \operatorname{sign}
    \left(\theta_{p}^{i,j,k}(\sigma)\right)
    =
    \operatorname{sign}
    \left(\theta_{q}^{i,j,k}\right).
    \label{eq:no_flip}
\end{equation}
Thus, the assignment preserves the orientation of the target formation and excludes unintended reflected realizations while allowing global translations and rotations.

\begin{figure}[!t]
    \centering
    \includegraphics[width=1\columnwidth]{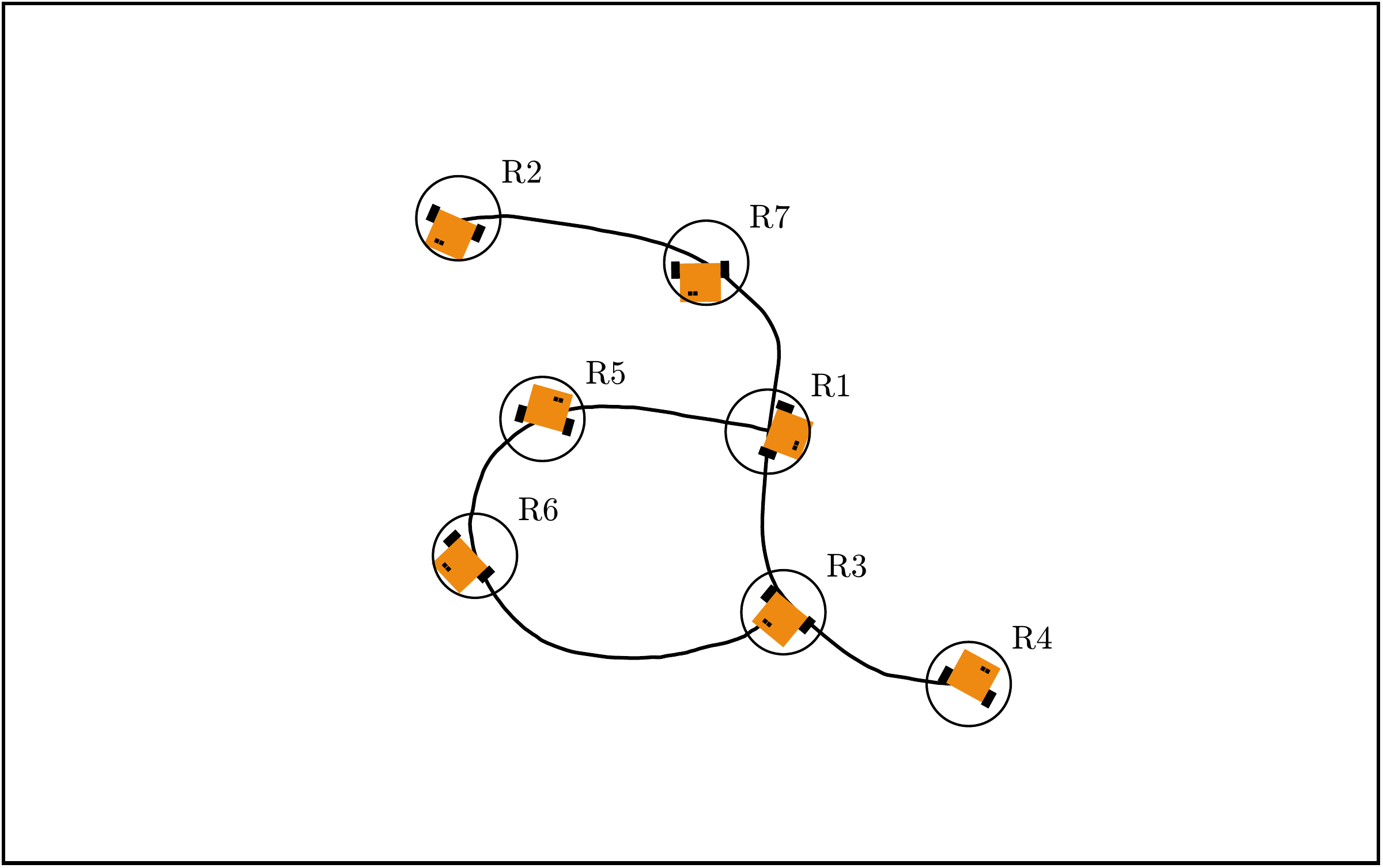}
    \caption{Robotarium~\cite{pickem2017robotarium} simulation of the example letter ``a” formation. Robots $R1$-$R7$ occupy the target positions obtained from the filtered target points in Figure~\ref{fig:critical_points}. The realized configuration preserves the geometry and orientation of the user-specified formation while allowing the global translation and rotation inherent to distance-based formation control.}
    \label{fig:letter_a_final}
\end{figure}

To quantify the translation and rotation between the target and realized configurations, let $\theta$ denote the rotation angle and $\boldsymbol{\tau}\in\mathbb{R}^2$ the translation vector aligning the target configuration $Q$ with the realized robot configuration $P$. These parameters describe the global rigid transformation permitted by the distance-based formation controller and are used to characterize the final formation realization. Figure~\ref{fig:letter_a_final} shows the resulting Robotarium realization of the same seven-node letter ``a'' example used in Figure~\ref{fig:letter_a_CV_pipeline}. For this realization, the alignment with the target configuration corresponds to $\theta=8.29^\circ$ and $\boldsymbol{\tau}=(-0.089,-0.105)$~m.

\section{User Study} \label{sec:user_study}

A user study\footnote{The study protocol was reviewed and approved by the University of California, Irvine Institutional Review Board (IRB) under protocol STUDY00000453.} was conducted to evaluate the usability of freehand sketching as an end-user programming interface for robot-swarm formation specification. In this section, we detail the user study procedure and participant selection as well as the experiment setup and the standardized questionnaire used to evaluate the study.

\subsection{Participants and Study Procedure} \label{ssec:participants}

Participants were recruited from the University of California, Irvine (UCI) community through an open recruitment form describing the study as an interactive experiment in which users would control a group of robots using drawings on a tablet. No prior robotics experience was required for participation. Twenty participants completed the study, comprising 12 undergraduate students, 6 graduate students, and 2 postdoctoral researchers. Participants represented several engineering and science backgrounds, including electrical, computer, mechanical, aerospace, mechatronics, and biological sciences.

\begin{figure}[!t]
    \centering
    \includegraphics[width=\columnwidth]{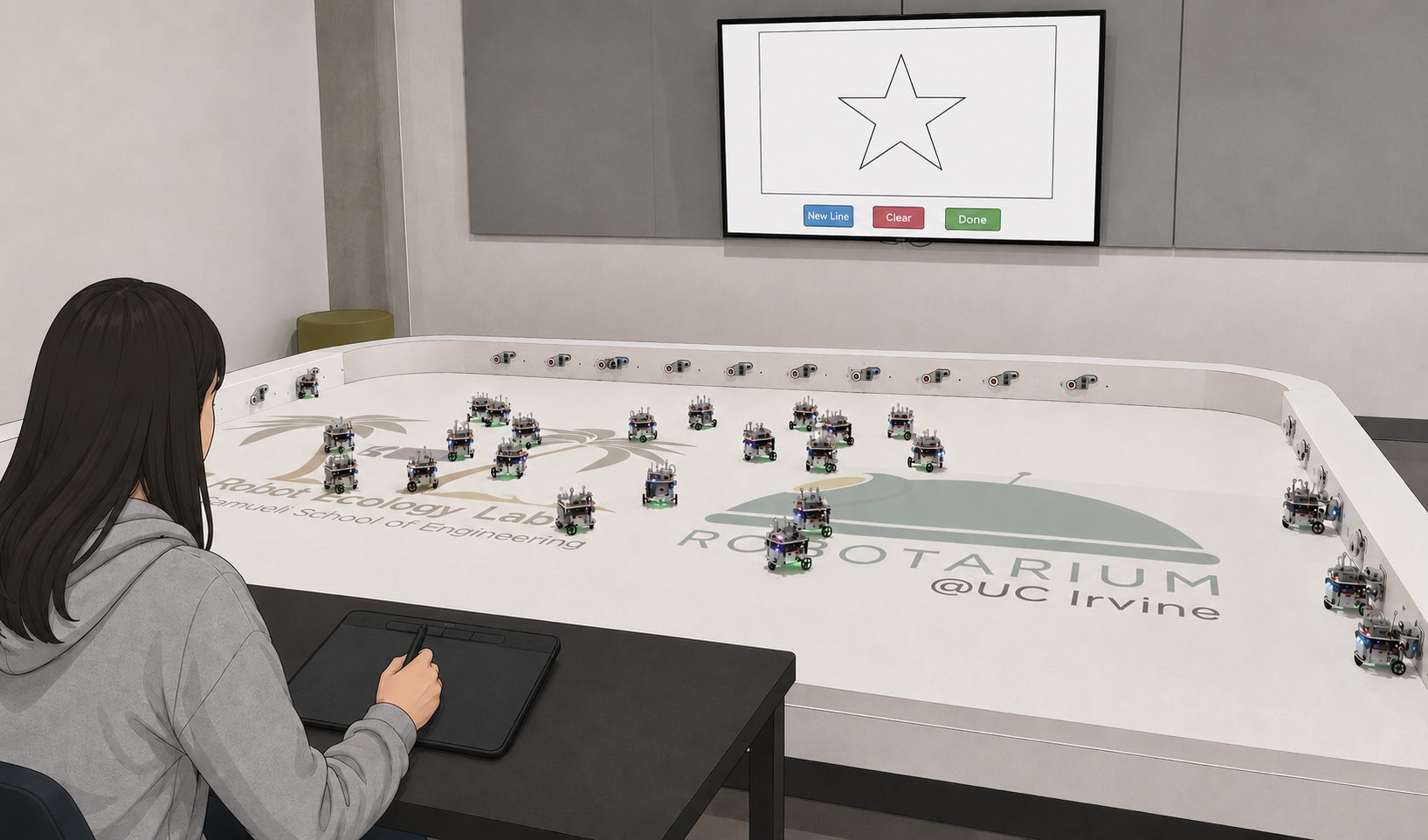}
    \caption{ChatGPT-generated illustration of the experimental setup used in the user study, based on the actual Robotarium environment and interface configuration. Participants specified desired swarm formations using a stylus on a wireless Wacom drawing tablet while viewing the GUI on the television positioned above the physical Robotarium testbed. The \emph{New Line}, \emph{Clear}, and \emph{Done} controls support drawing, resetting, and submitting the sketch, respectively.}
    \label{fig:user_study_setup}
\end{figure}

At the beginning of each session, participants were presented with a short instruction window before gaining access to the graphical user interface (GUI). The instructions described only the basic interface operations: pressing \emph{New Line} to draw, \emph{Clear} to reset the drawing canvas, and \emph{Done} to submit the completed drawing. No example or prescribed formation was provided as part of the instructions.

Participants were free to choose the formations they wanted the robot swarm to establish, with no prescribed shapes or restrictions on the semantic content of their drawings. Each participant was required to complete at least one sketch-to-swarm execution and could subsequently submit additional drawings if desired. This open-ended protocol allowed the interface to be evaluated using self-selected spatial specifications.

For each submitted drawing, the sketch was processed through the pipeline described in Section~\ref{sec:framework}, after which participants directly observed the physical robot team realize the resulting formation. After interacting with the system, participants completed the System Usability Scale (SUS) questionnaire described in Section~\ref{sec:sus}. Each experimental session, including interaction with the system and completion of the questionnaire, lasted approximately 30 minutes or less. The Robotarium experiments were also recorded, with no participant faces captured in the recordings.

\subsection{Experiment Setup \& Graphical User Interface} \label{sec:experiment_setup_and_GUI}

The user study was conducted using the physical Robotarium testbed in the UCI Robot Ecology Lab. As illustrated in Figure~\ref{fig:user_study_setup}, participants interacted with the sketch-to-swarm system using a Wacom drawing tablet connected via Bluetooth to the host computer. The GUI was displayed on a television positioned in front of the Robotarium testbed, allowing participants to draw on the tablet while viewing the interface and subsequently observe the robot team physically realize their input within the same experimental setup. This arrangement provided a direct interaction loop between the participant's freehand specification and its physical realization by the swarm. The GUI consists primarily of a blank drawing canvas whose dimensions are proportional to those of the Robotarium workspace, as well as three buttons, as shown in Figure~\ref{fig:user_study_setup}.

The physical experiments were performed using up to 20 mobile robots operating within the Robotarium testbed. Because the number of extracted formation nodes cannot exceed the number of robots available for physical deployment, the sketch-processing stage constrains and filters the extracted critical points accordingly. The resulting target configuration is then assigned to the available robots and realized using the distributed formation controller described in Section~\ref{ssec:formation_control}.

\subsection{System Usability Scale} \label{sec:sus}

Following interaction with the system, participants completed the System Usability Scale (SUS)~\cite{brooke1996sus}, a standardized ten-item questionnaire for assessing perceived system usability. Each statement is rated on a five-point Likert scale from 1 (\emph{strongly disagree}) to 5 (\emph{strongly agree}). The questionnaire alternates between positively and negatively worded statements and provides a compact assessment of users' overall perception of the system, including aspects related to ease of use, complexity, learnability, integration, and confidence.

Following the standard SUS scoring procedure, the contribution of each positively worded odd-numbered item was computed by subtracting one from its response, while the contribution of each negatively worded even-numbered item was computed by subtracting its response from five. The ten adjusted contributions were then summed and multiplied by 2.5, yielding an overall score between 0 and 100 for each participant:
\begin{equation}
    S_{\mathrm{SUS}}=2.5
    \left[
    \sum_{i\in\{1,3,5,7,9\}}(r_i-1)
    +
    \sum_{i\in\{2,4,6,8,10\}}(5-r_i)
    \right],
    \label{eq:sus}
\end{equation}
where $r_i\in\{1,\ldots,5\}$ denotes the response to item $i$. SUS scores are not percentages, but standardized usability scores intended for comparison against established empirical benchmarks. A score of approximately 68 is commonly used as a reference for average usability~\cite{bangor2008empirical}.

\section{Results} \label{sec:results}

The results of the user study described in Section~\ref{sec:user_study} are evaluated from two complementary perspectives: participants' perceived usability of the sketch-based interface and the physical realization of their user-specified formations on the Robotarium. Together, these results characterize both the end-user experience and the system's ability to translate freehand specifications into physical swarm formations.

\subsection{User Study Results and Analysis} \label{ssec:user_study_results}

\begin{figure}[!b]
    \centering
    \includegraphics[width=\columnwidth]{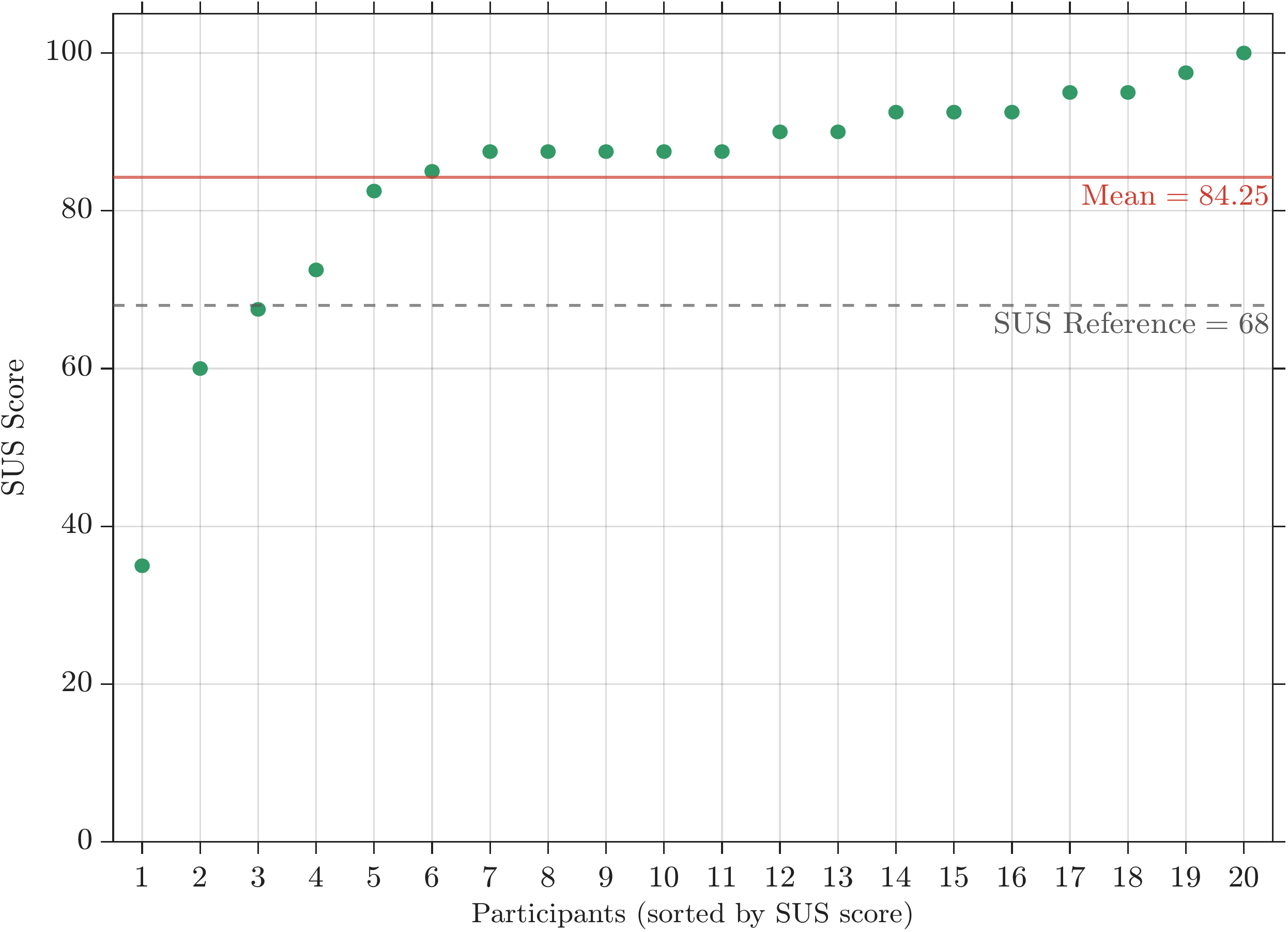}
    \caption{Participant-level System Usability Scale (SUS) scores, sorted in ascending order. The dashed line indicates the commonly reported SUS reference value of 68~\cite{bangor2008empirical}, while the solid line indicates the study mean of 84.25. Seventeen of the 20 participants scored above the reference value.}
    \label{fig:sus_participant_scores}
\end{figure}

All 20 participants who completed the physical interaction study also submitted the post-study SUS questionnaire. The participant-level SUS scores had a mean of $84.25$ ($SD=15.24$) and a median of $87.5$, with scores ranging from $35$ to $100$, as shown in Figure~\ref{fig:sus_participant_scores}. The observed mean is substantially above the commonly reported SUS reference value of 68~\cite{bangor2008empirical}, indicating high perceived usability of the sketch-to-swarm interface. Seventeen of the 20 participants obtained SUS scores above 68, and 16 obtained scores of at least 80.

\begin{table}[!t]
    \centering
    \caption{Participant responses to the ten System Usability Scale items. Values report the mean and standard deviation of the original 1--5 Likert responses across the 20 participants.}
    \label{tab:sus_items}
    \resizebox{\columnwidth}{!}{
    \begin{tabular}{clcc}
        \hline
        \textbf{Item} & \textbf{SUS statement} & \textbf{Mean} & \textbf{SD} \\
        \hline
        1 & Would like to use the system frequently & 4.05 & 0.89 \\
        2 & System was unnecessarily complex & 1.65 & 0.99 \\
        3 & System was easy to use & 4.30 & 0.86 \\
        4 & Would need support of a technical person & 1.95 & 1.15 \\
        5 & Functions were well integrated & 4.40 & 0.82 \\
        6 & System contained too much inconsistency & 1.55 & 0.69 \\
        7 & Most people would learn the system quickly & 4.55 & 0.76 \\
        8 & System was cumbersome to use & 1.50 & 0.83 \\
        9 & Felt confident using the system & 4.40 & 0.82 \\
        10 & Needed to learn a lot before using the system & 1.35 & 0.59 \\
        \hline
    \end{tabular}}
\end{table}

\begin{figure*}[!t]
    \centering
    \subfloat[Participant Sketch 1.]{
        \fbox{\includegraphics[width=0.235\textwidth]{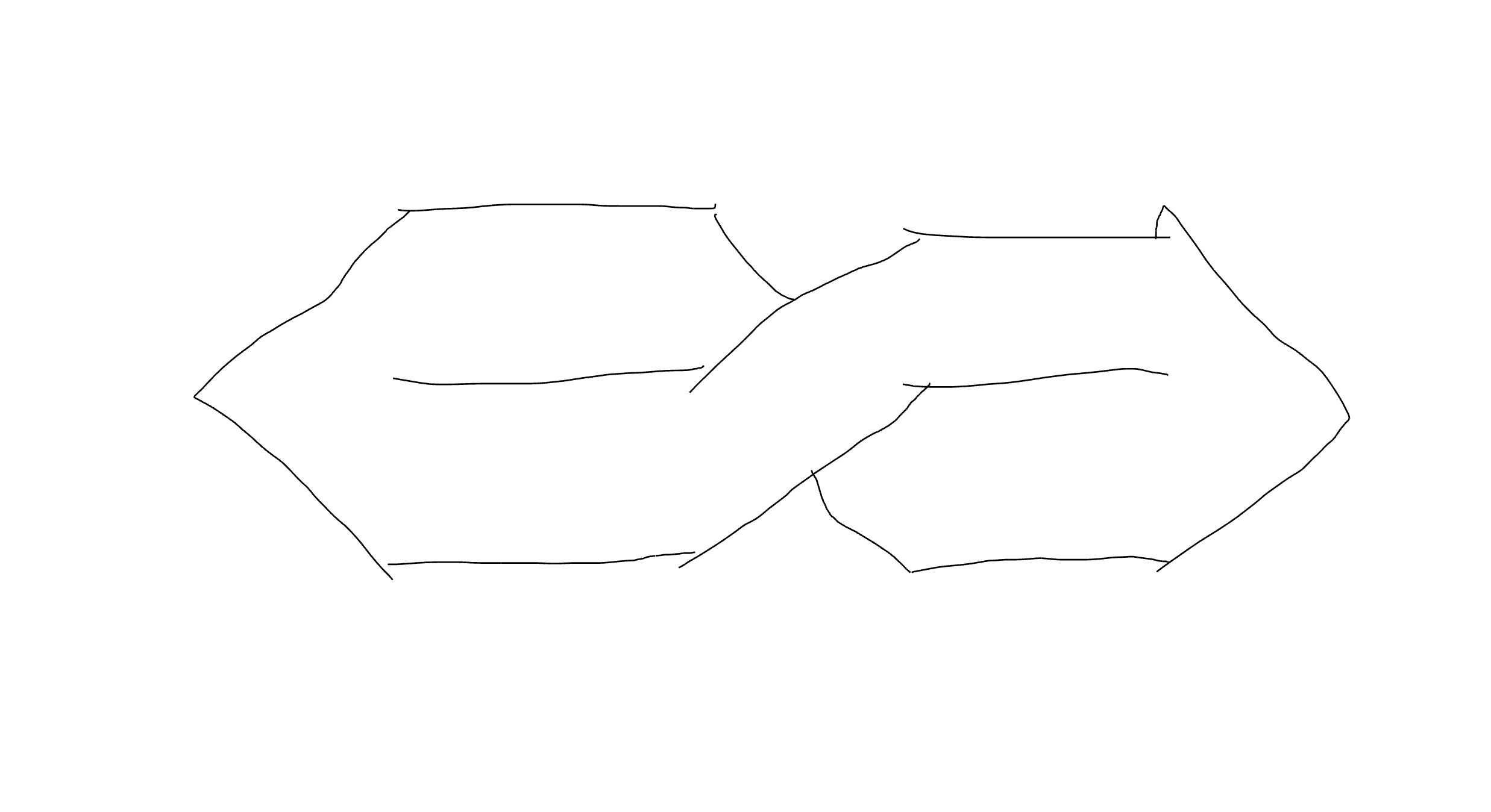}}
        \label{fig:user1_sketch}
    }
    \hfill
    \subfloat[Converged Formation 1.]{
        \includegraphics[width=0.235\textwidth]{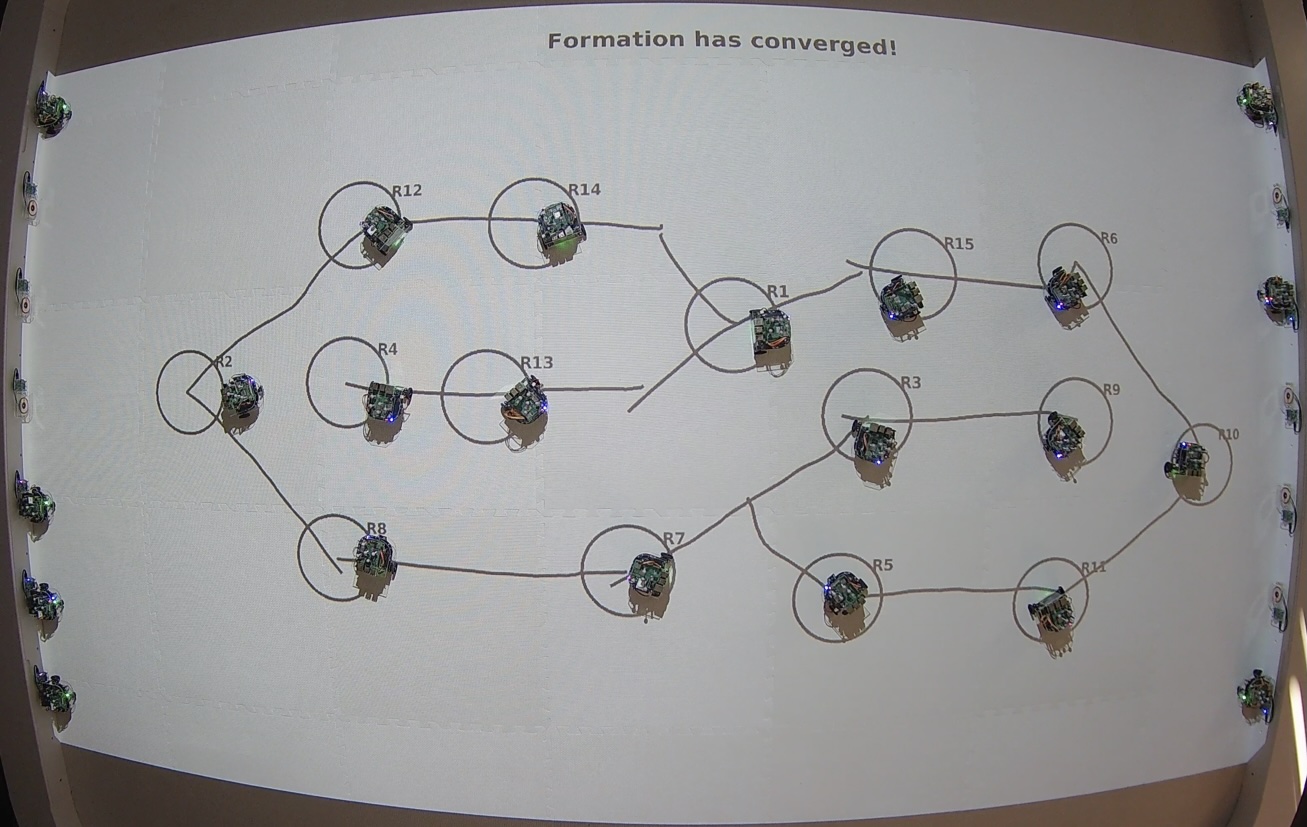}
        \label{fig:user1_robotarium}
    }
    \hfill
    \subfloat[Participant Sketch 2.]{
        \fbox{\includegraphics[width=0.235\textwidth]{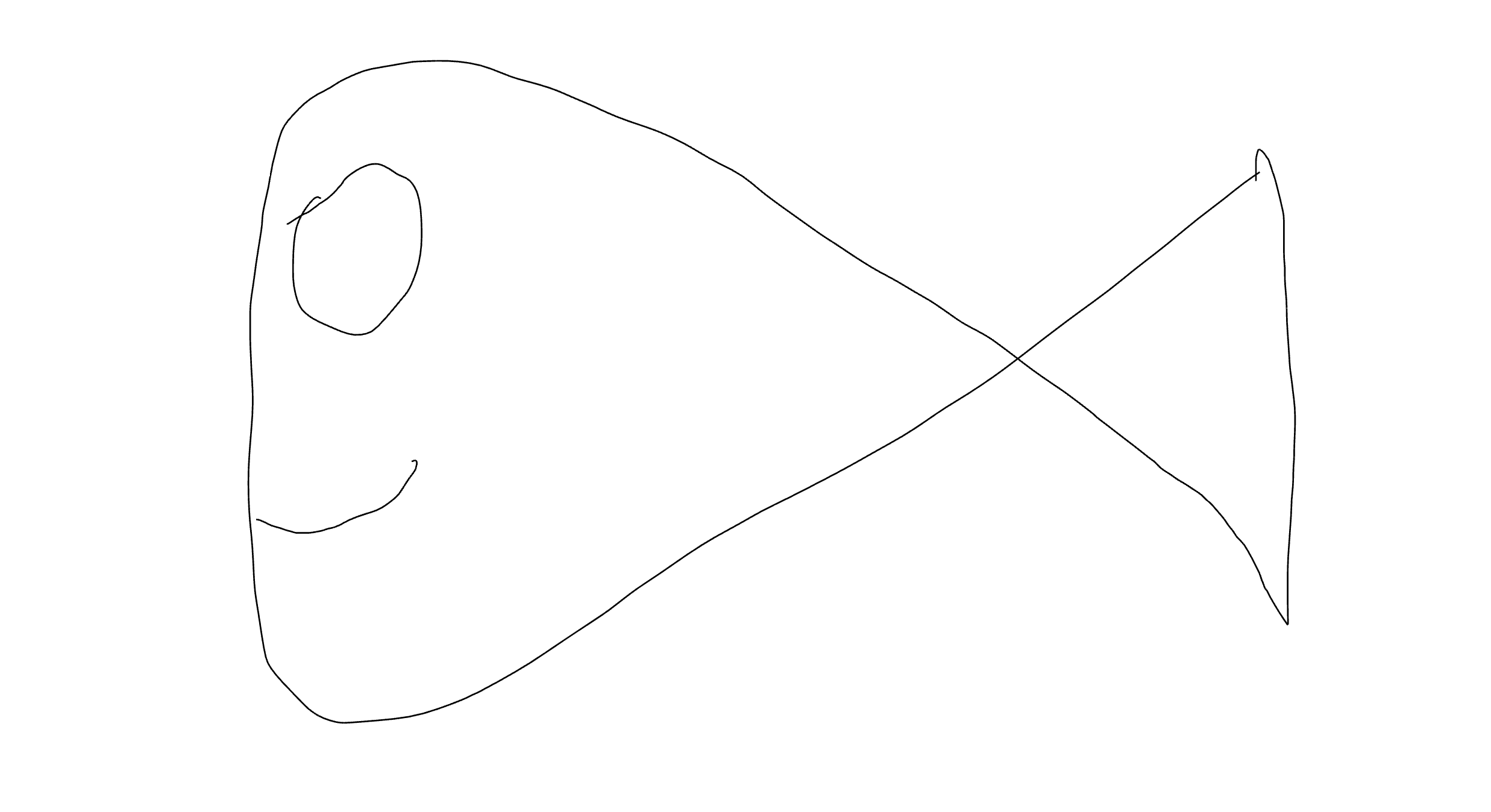}}
        \label{fig:user2_sketch}
    }
    \hfill
    \subfloat[Converged Formation 2.]{
        \includegraphics[width=0.235\textwidth]{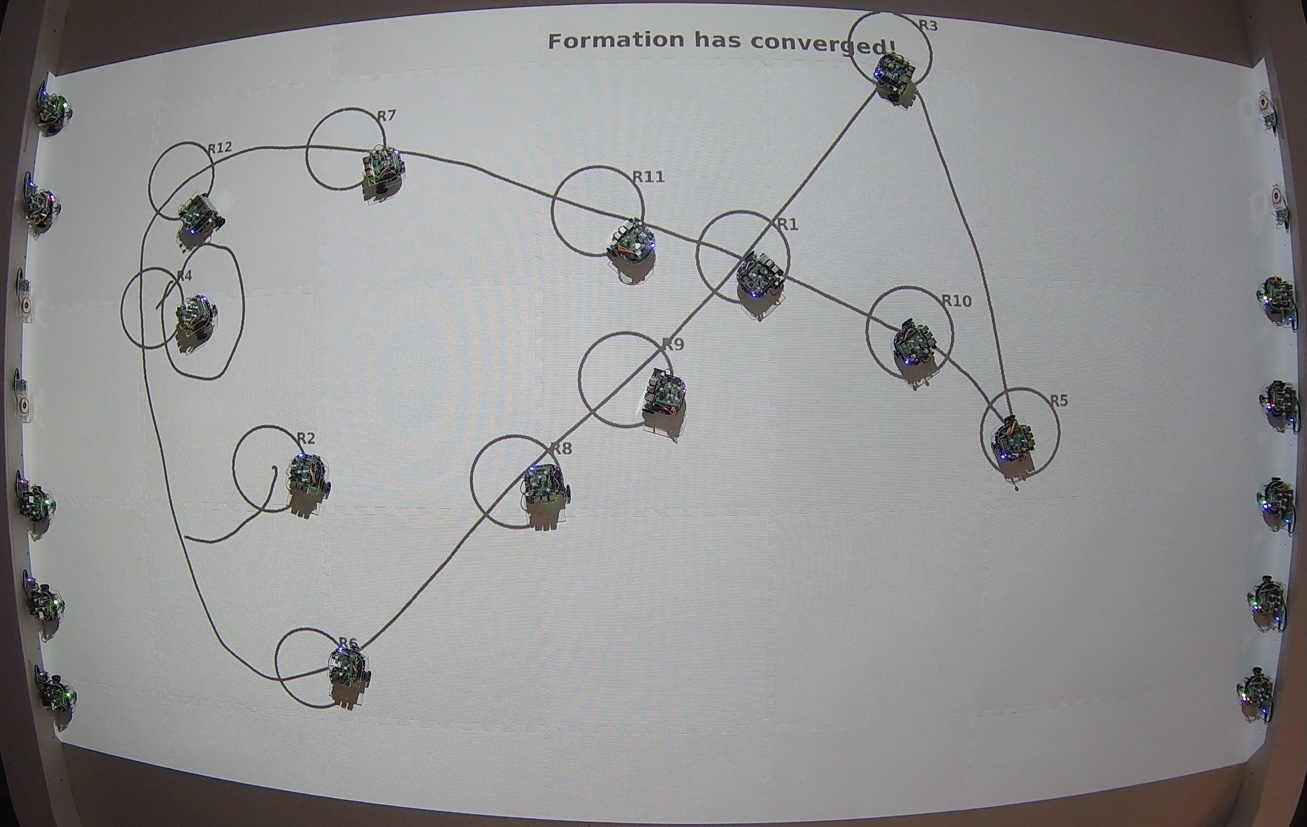}
        \label{fig:user2_robotarium}
    }
    \hfill
    \subfloat[Participant Sketch 3.]{
        \fbox{\includegraphics[width=0.235\textwidth]{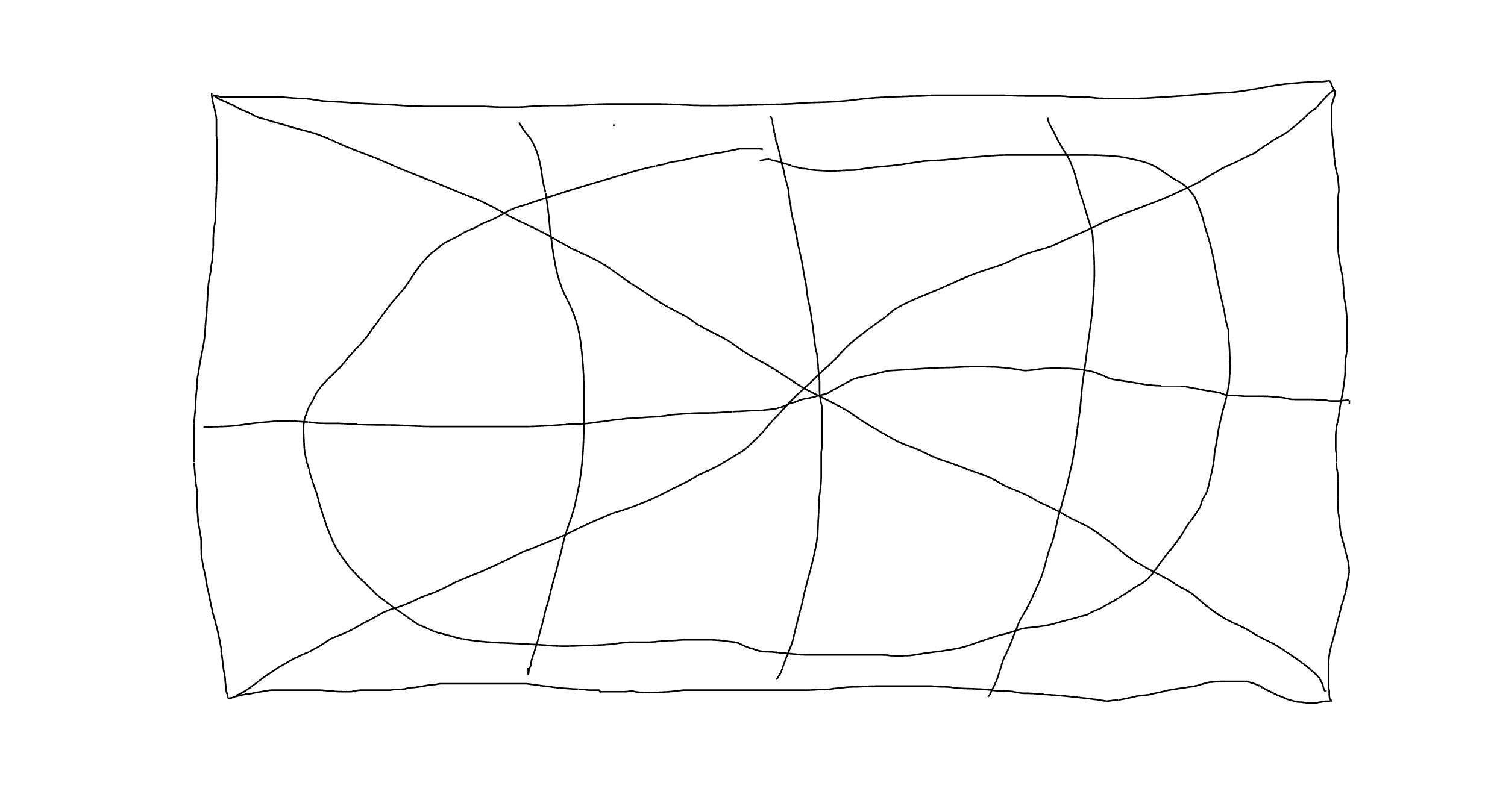}}
        \label{fig:user3_sketch}
    }
    \hfill
    \subfloat[Converged Formation 3.]{
        \includegraphics[width=0.235\textwidth]{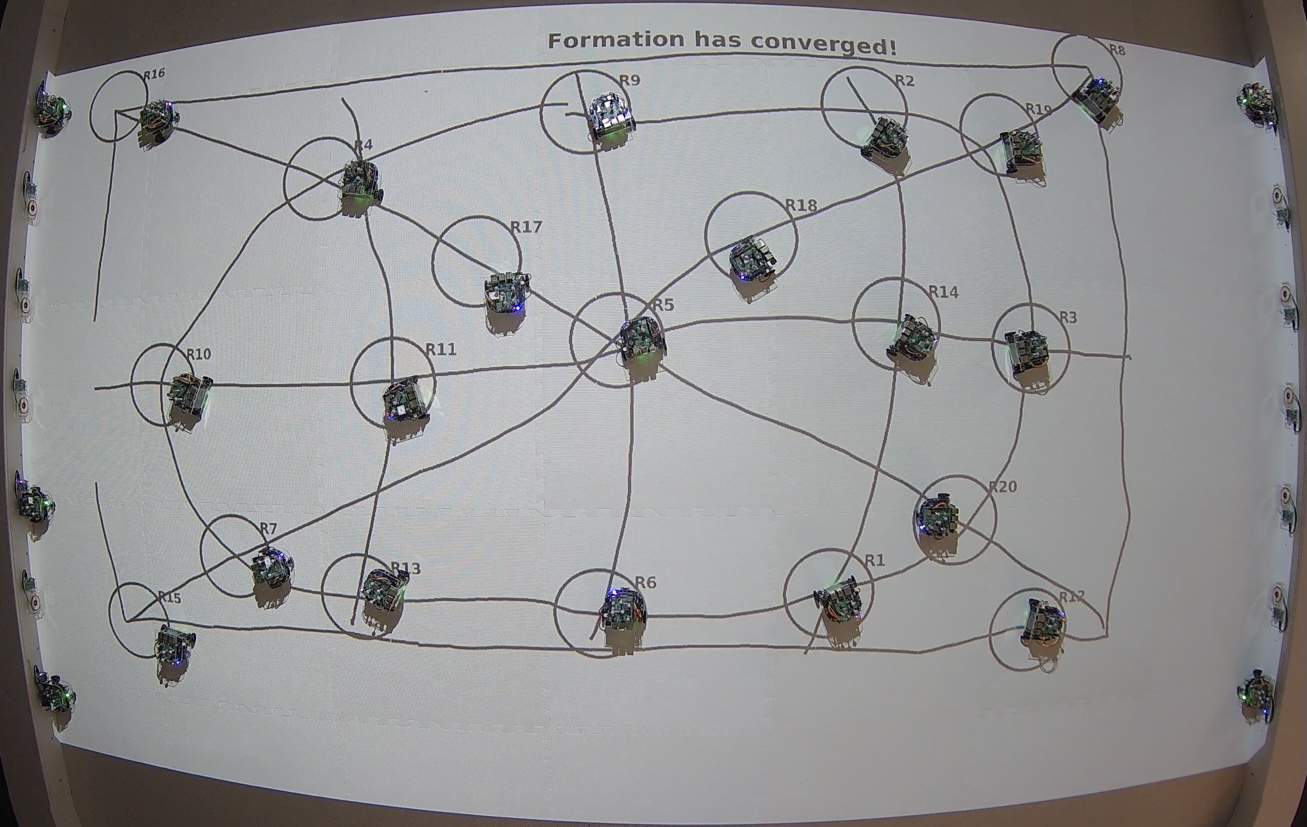}
        \label{fig:user3_robotarium}
    }
    \hfill
    \subfloat[Participant Sketch 4.]{
        \fbox{\includegraphics[width=0.235\textwidth]{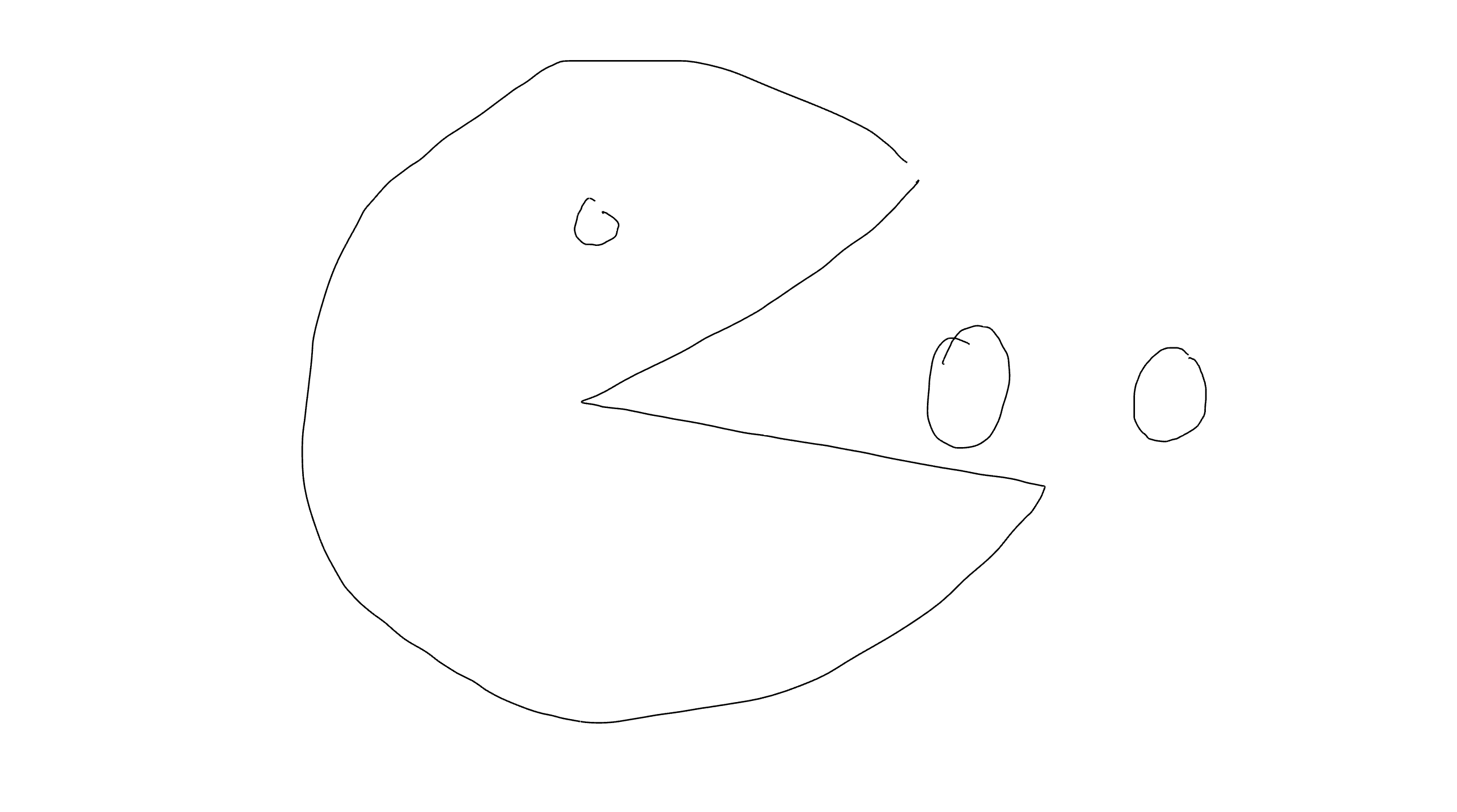}}
        \label{fig:user4_sketch}
    }
    \hfill
    \subfloat[Converged Formation 4.]{
        \includegraphics[width=0.235\textwidth]{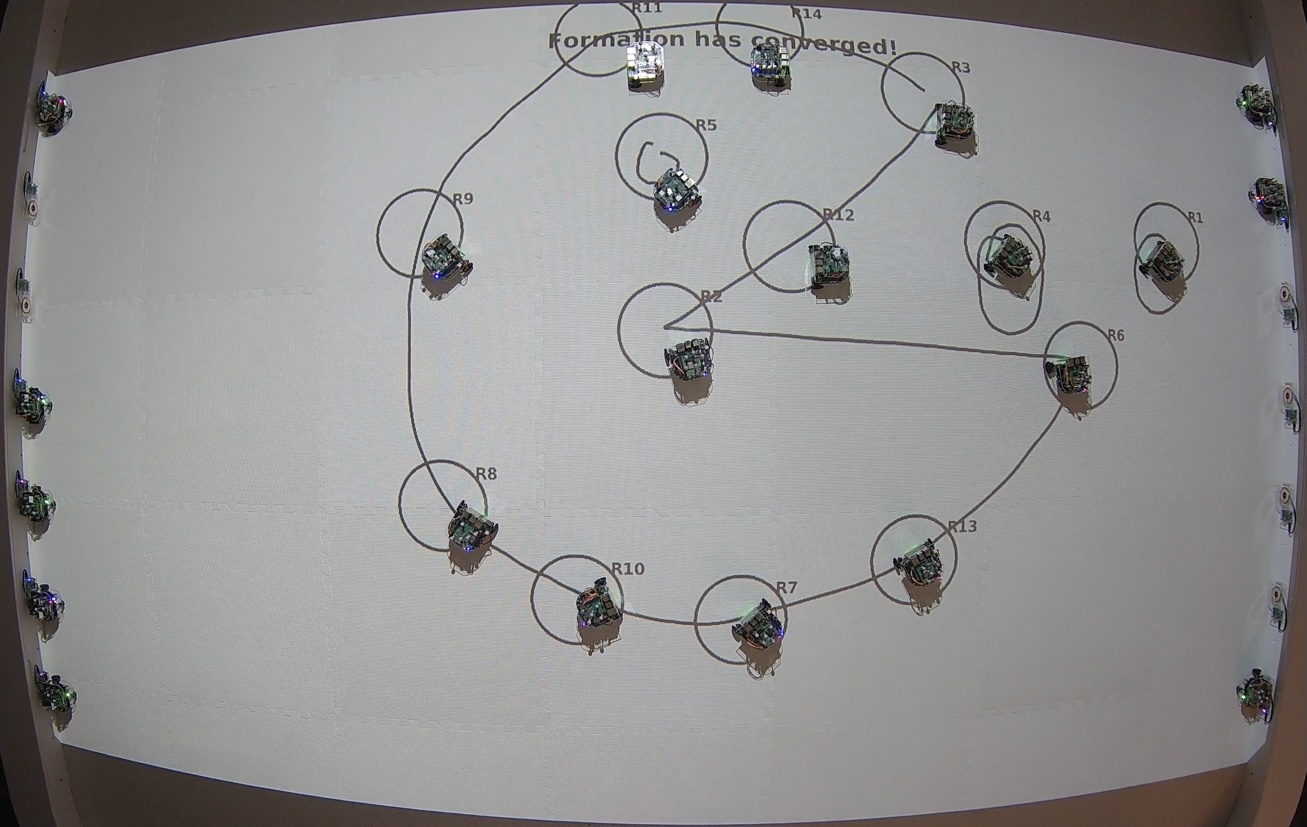}
        \label{fig:user4_robotarium}
    }
    \hfill
    \subfloat[Participant Sketch 5.]{
        \fbox{\includegraphics[width=0.235\textwidth]{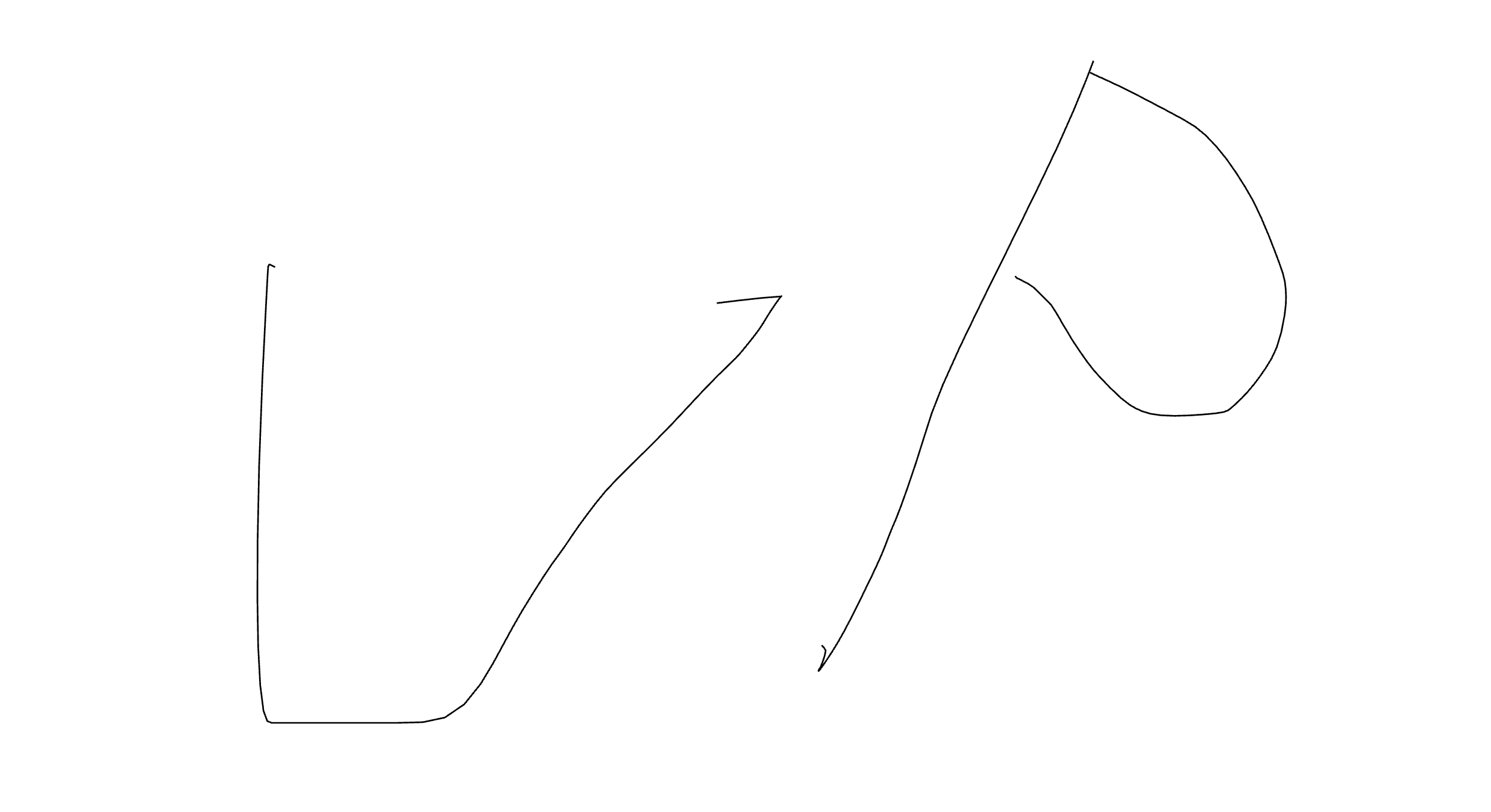}}
        \label{fig:user5_sketch}
    }
    \hfill
    \subfloat[Converged Formation 5.]{
        \includegraphics[width=0.235\textwidth]{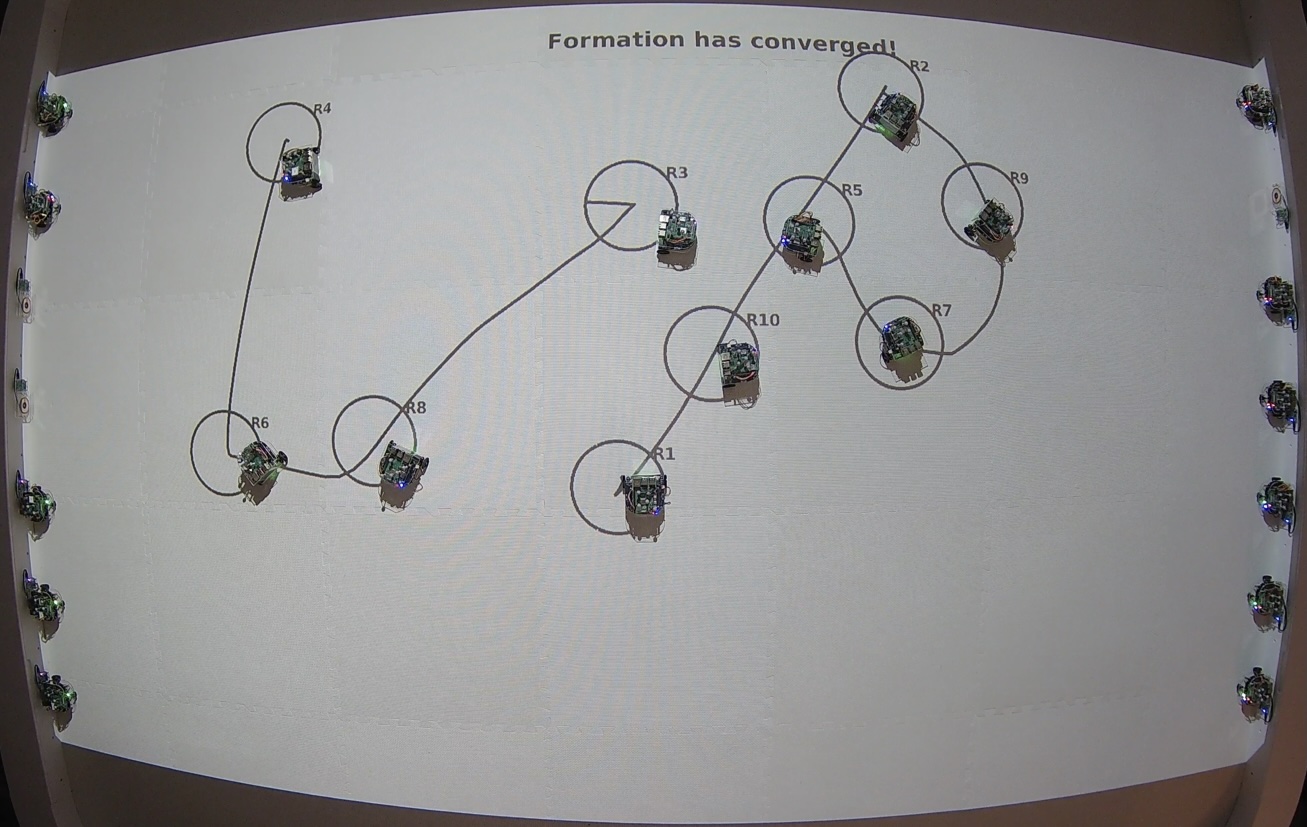}
        \label{fig:user5_robotarium}
    }
    \hfill
    \subfloat[Participant Sketch 6.]{
        \fbox{\includegraphics[width=0.235\textwidth]{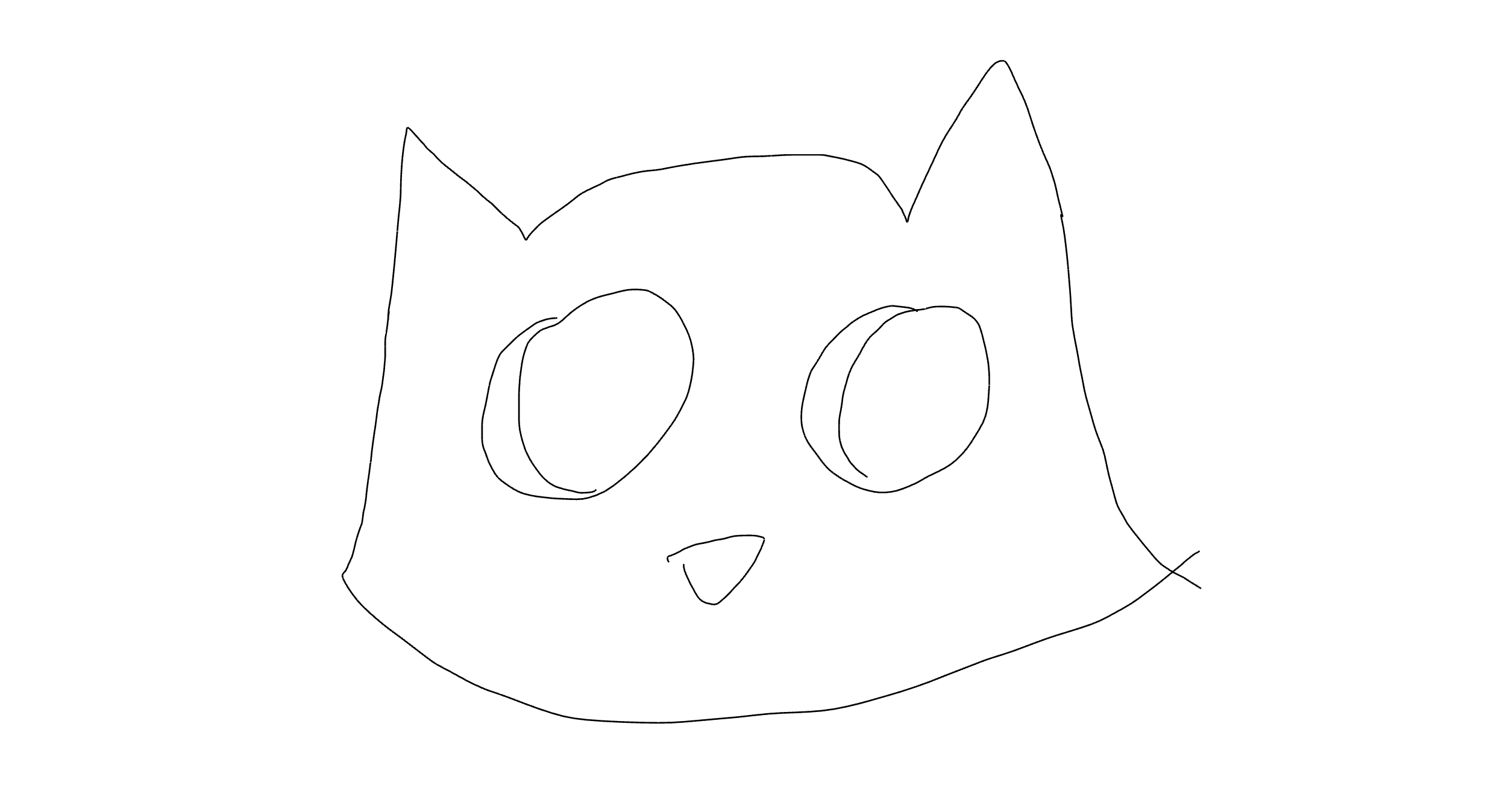}}
        \label{fig:user6_sketch}
    }
    \hfill
    \subfloat[Converged Formation 6.]{
        \includegraphics[width=0.235\textwidth]{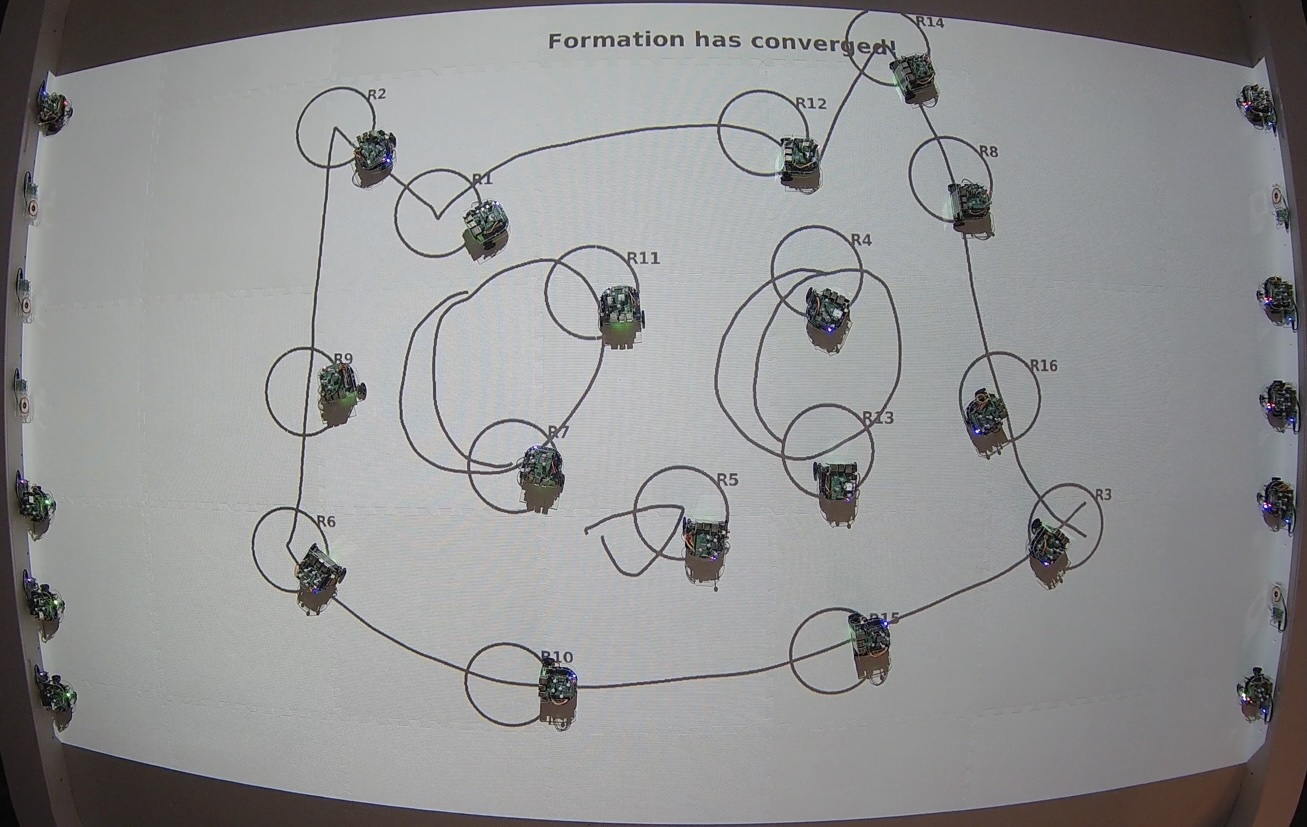}
        \label{fig:user6_robotarium}
    }
    \caption{Six of the 42 participant-generated formations from the physical user study. Each sketch/formation pair shows a freehand sketch and the corresponding converged formation executed on the physical Robotarium testbed. The six examples use 15, 12, 20, 14, 10, and 16 robots, respectively. The realized formations may undergo a global rotation and translation relative to the original sketches, as permitted by the distance-based formation controller, while preserving their orientation; no reflected realizations were observed.}
    \label{fig:user_formation_examples}
\end{figure*}

The item-level responses in Table~\ref{tab:sus_items} provide additional insight into the aggregate SUS result. The strongest response was obtained for Item~7, for which participants rated the statement that most people would learn the system quickly at $4.55/5$. This result is complemented by Item~10, for which participants gave a mean response of only $1.35/5$ to the statement that they needed to learn a substantial amount before using the system. Participants also reported high confidence while interacting with the system (Item~9, $4.40/5$), perceived its functions as well integrated (Item~5, $4.40/5$), and rated it as easy to use (Item~3, $4.30/5$). Conversely, participants generally disagreed that the system was cumbersome (Item~8, $1.50/5$), inconsistent (Item~6, $1.55/5$), or unnecessarily complex (Item~2, $1.65/5$). These responses indicate favorable perceptions across multiple aspects of usability and learnability.

In addition to the questionnaire responses, the open-ended study protocol resulted in repeated interaction with the system. Across the 20 participants, 42 self-selected drawings were submitted and physically executed, corresponding to $2.10\pm1.02$ formations per participant. Seven participants completed one formation, six completed two, five completed three, and two completed four. Thus, 65\% of participants chose to perform more than the single required execution. Because additional executions were optional, these counts are reported as descriptive observations rather than as a formal measure of engagement.

A formation was considered converged when the absolute inter-agent distance error for every prescribed edge $(i,j)\in\mathcal{E}$ satisfied
\begin{equation}
    \left|
    \|\mathbf{p}_i-\mathbf{p}_j\|_2-d_{ij}
    \right|
    \leq 0.025\mathrm{m}.
    \label{eq:convergence_tolerance}
\end{equation}
All 42 formations specified in the user study satisfied this convergence criterion.

Taken together, the results indicate that participants generally perceived freehand sketching as a usable and readily learnable interaction abstraction. These findings are particularly relevant given that participants received only basic GUI instructions and were neither trained to construct particular formations nor provided predefined shapes to reproduce. The physical realization of representative participant-generated formations is examined further in Section~\ref{ssec:physical_results}.

Some variability in user experience nevertheless remains. Item~4, concerning whether technical assistance would be required to use the system, exhibited the largest variation ($1.95\pm1.15$). The participant-level distribution in Figure~\ref{fig:sus_participant_scores} also shows that, despite generally high SUS scores, perceived usability was not uniform across all participants. Thus, although the aggregate evaluation indicates high perceived usability, the interface was not equally intuitive for all participants and may benefit from further refinement of its interaction cues and introductory guidance.

The results should also be interpreted within the scope of the study. Participants were primarily university students and researchers, many from engineering disciplines, and therefore do not represent the broader population of non-technical users. The study intentionally allowed participants to select both their drawings and the number of interactions rather than imposing standardized formation tasks. Consequently, the experiment evaluates perceived usability under open-ended interaction rather than comparative task performance across participants.

\subsection{Physical Robotarium Formation Results}
\label{ssec:physical_results}

The physical experiments also show the range of spatial configurations participants chose to specify through the interface. Figure~\ref{fig:user_formation_examples} presents examples from the user study, pairing participant sketches with the corresponding converged formations on the physical Robotarium testbed. Each sketch was processed independently through the pipeline in Section~\ref{sec:framework}. The examples demonstrate that visually distinct freehand inputs can be translated into physical multi-robot configurations while preserving the intended overall geometry and orientation of the user-specified formation.

Although the number and complexity of the resulting formation points depend on the geometry of the input sketch and the number of robots available for execution, the participants were not required to account for these constraints explicitly. Instead, these considerations were handled by the system after the sketch was submitted. The physical realizations therefore further illustrate the division of roles underlying the proposed human-swarm collaboration: the human specifies the desired shape, while the swarm determines how that specification is represented and physically achieved.

\section{Conclusion} \label{sec:conclusion}

This paper investigated freehand sketching as an end-user programming abstraction for human-swarm formation specification. The user study supports the potential of this interaction paradigm for accessible human-swarm collaboration without requiring robotics or programming expertise. The present evaluation was limited to a primarily university-based participant population and static planar formations. Future work will consider broader non-technical user populations and extend sketch-based interaction to swarm behaviors beyond formation control, as well as continuous human-in-the-loop integration.

\section*{Acknowledgments}

The authors would like to thank Dr. Alexander Nguyen for helpful discussions.

\bibliographystyle{IEEEtran}
\bibliography{mybib}

\end{document}